\documentclass{article}
\usepackage{spconf,amsmath,graphicx,hyperref}

\usepackage{cite}
\usepackage{graphicx}
\usepackage{booktabs}
\usepackage[caption=false,font=footnotesize]{subfig}
\usepackage{xcolor}
\usepackage{tikz}
\usetikzlibrary{spy}

\usepackage{amsmath,amssymb,amsfonts}
\usepackage{algorithmic}
\usepackage{graphicx}
\usepackage{textcomp}

\usepackage{subcaption} 
\usepackage{multirow}  
\usepackage{subcaption}
\usepackage{pgfplots}
\pgfplotsset{compat=1.18}   
\usepackage{indentfirst}

\title{YOU'VE SEEN ENOUGH: QUALITY-CONSTRAINED IMAGE CODING FOR MACHINES}
\name{\begin{tabular}[t]{@{}c@{}}Khoa Pham-Dinh$^1$, Sanaz Nami$^1$\\
Hamed Rezazadegan Tavakoli$^2$, Moncef Gabbouj$^1$, Farhad Pakdaman$^2$\end{tabular}}
\address{Tampere University$^1$, Nokia Technologies$^2$}

\begin{document}
\ninept
\maketitle
\begin{abstract}
Visual data is increasingly consumed by machine-vision systems rather than by human observers. Image Coding for Machines (ICM) compresses images assuming the main observer is a computer vision application and that the human observer needs to inspect or validate the decisions. Inspired by just-noticeable distortion, we cap human-observed quality at a desired level and devote the remaining bits to machine performance. Specifically, joint compression-segmentation training is recast as a constrained optimization problem in which the codec must meet a predefined acceptable target visual quality while a task term consumes the remaining coding capacity. This paper proposes two variants of a penalty function that guides the quality toward the target: an absolute function and a bilinear function, the latter applying a steeper slope once the target visual quality is exceeded. Experimental results show that, under the quality constraint, the proposed method achieves BD-rates of $-22.82\%$ and $-29.81\%$ relative to an unconstrained joint rate--distortion--task optimization and a simple rate--distortion baseline, respectively, showcasing bitrate reduction with the same task performance. This is achieved while the codec also meets the target visual quality with a reasonable error and without adding any complexity overhead.
\end{abstract}
%

\section{Introduction}
Image and video data dominate global network traffic, yet a substantial part of modern visual content is never consumed by a human; it is processed directly by machine-vision systems for computer vision tasks such as object detection, classification, and semantic segmentation. Classical codecs such as AVC, HEVC, and VVC rely on handcrafted tools that exploit the human visual system. Most end-to-end learned image compression (LIC) methods replace the hand-crafted pipeline of classical codecs with
neural analysis and synthesis networks, trained on a relaxed rate--distortion
objective $L=R+\lambda D$~\cite{balleEndtoendOptimizedImage2017,minnenJointAutoregressiveHierarchical2018,chengLearnedImageCompression2020,heELICEfficientLearned2022,yangLossyImageCompression2023,heCheckerboardContextModel2021,koyuncuEfficientContextformerSpatioChannel2024,pakdamanChannelWiseFeatureDecorrelation2024,zhangGeneralizedGaussianModel2025}. Conventionally, LIC is optimized for human perception, with a focus on metrics such as Peak Signal-to-Noise Ratio (PSNR) and Multiscale Structural Similarity (MS-SSIM). These metrics, while designed to preserve fidelity, correlate weakly with the accuracy of the vision models that ultimately process the decoded images, so bits are routinely spent on texture details that are task irrelevant, but task-critical structure may be lost during reconstruction. As a result, the need to design machine-centric codecs that preserve semantic information rather than pixel fidelity has emerged.

In recent years, ICM has emerged as a promising research direction, driven by the increasing reliance on automated machine analysis for image interpretation. Machine-targeted compression algorithms aim to reduce the bitrate while maintaining high performance on machine tasks such as classification, detection, and segmentation. Representative work can be broadly classified into two main paradigms~\cite{choiCompressAIVisionOpensourceSoftware2025,eimonEmergingStandardsMachinetoMachine2026}, following the standardization efforts of the Moving Picture Experts Group (MPEG). The first, Feature Coding for Machines (FCM), targets machine-only consumption, where no human observes the decoded stream. These methods compress intermediate features rather than image pixels~\cite{duanEfficientFeatureCompression2022,fengImageCodingMachines2022}. The second paradigm, Video/Image Coding for Machines (VCM/ICM), operates in the pixel domain and jointly optimizes codecs for hybrid human-machine consumption, where the content must remain interpretable to a human who may inspect or supervise the decisions. This paradigm preserves a decodable image and trains the codec with an additional task loss for joint human-machine consumption. Recent standardization efforts in VCM/ICM have attracted considerable attention, encompassing the development of coding tools and support for diverse real-world use cases~\cite{gaoRecentStandardDevelopment2021,duanVideoCodingMachines2020,ascensoJPEGAIStandard2023,pham-dinhEvaluatingEmergingMPEG2025}. While some works improve machine performance by pre- or post-processing in the image domain~\cite{stankiewiczRegionofInterestBasedVideoCoding2024,luPreprocessingEnhancedImage2024}, progress in VCM/ICM has largely stemmed from advances in adapting LIC models for machine consumption~\cite{leImageCodingMachines2021,leLearnedImageCoding2021,choiScalableImageCoding2022,hePOELICPerceptionOrientedEfficient2022,ahonenNNVVCHybridLearnedConventional2024,shindoImageCodingMachines2024}. We target an ICM setting similar to~\cite{leImageCodingMachines2021}. Unlike these methods, which tune the loss weighting to trade off quality and task performance, we constrain the distortion to an explicit target.

Motivated by just-noticeable-difference (JND) studies, which indicate that a human observer is satisfied once reconstruction passes a ``good-enough'' quality~\cite{namiBLJUNIPERCNNAssistedFramework2023}, we argue that fidelity beyond that level wastes bits that could instead serve the task. In a joint objective optimization scheme, the balance between quality and task is governed only implicitly by using two Lagrangian multipliers; that is to say, the quality realized at a given bitrate is whatever the optimizer settles on, rather than being guided to a desired level. This raises the central question of this work: \emph{how to control a codec's output quality while reallocating the learning capacity freed from perceptual fidelity toward the downstream task?}

We propose a training scheme that treats the target quality as an explicit constraint. Training is formulated as a constrained optimization problem that minimizes the rate and task losses subject to the distortion matching a user-defined target $D_t$. This constraint is enforced through a linear penalty on the deviation of the distortion from $D_t$, yielding an unconstrained loss. Unlike a conventional rate--distortion objective, which always rewards lower distortion, the proposed loss penalizes any improvement beyond the target. We hypothesize that capping quality redirects the capacity otherwise spent on surplus quality toward task performance. Our contributions are:

\begin{itemize}
\item A quality-constrained rate--distortion--task formulation that enforces a
  target quality through a penalty term. The proposed training scheme reduces the bitrate required for segmentation while meeting the target quality.
\item Qualitative and quantitative evaluations on the Cityscapes semantic segmentation dataset, including an in-depth analysis of the coding gain and task performance.
\end{itemize}

\section{Related Work}

The prevailing training objective in learned image compression is the weighted rate--distortion loss $R + \lambda D$~\cite{balleEndtoendOptimizedImage2017}. Despite its efficacy, it is ill-suited for targeting specific operating points: the rate and quality reached for a given $\lambda$ depend on the model, and each image yields a slightly different bitrate or quality level, making rate matching laborious and imprecise. Alternative formulations have been proposed to control the operating point more directly. Toderici et al.~\cite{todericiVariableRateImage2016} proposed a recurrent architecture in which the bitrate is set by the number of encoding iterations, allowing a single model to operate at multiple rates. van Rozendaal et al.~\cite{rozendaalLossyCompressionDistortion2020} replaced the fixed tradeoff with constrained optimization, minimizing the rate subject to an upper bound on distortion. This allows a specific distortion value to be targeted without extensive per-model tuning of the tradeoff parameter, and enables pointwise model comparison, since models trained to the same distortion target can be compared by bitrate alone. Their method achieves rate--distortion performance comparable to that of the conventional weighted objective. However, the constraint is one-sided and involves no task objective, so quality beyond the target is never discouraged.

Several works have addressed rate targeting in learned image compression. For example, Guerin et al.~\cite{guerinRateconstrainedLearningbasedImage2022} recast the rate constraint as a penalty term, $\beta \left(\frac{R - R_t}{R_t}\right)^2$, that discourages deviations from a target bitrate $R_t$, while Xue et al.~\cite{xueLambdaDomainRateControl2024} predict, at encoding time, the $\lambda$ that yields the target bitrate using a lightweight estimator and an exponential $R$--$\lambda$ model, enabling single-pass encoding. These methods, however, target the bitrate rather than reconstruction quality and do not consider a machine-vision task.

Xie et al.~\cite{xieLastByteLearning2026} propose MVRNet, a no-reference JRD prediction framework that generates a spatially-varying quantization map by identifying how much compression each region can tolerate without degrading detection accuracy. By allocating more bits to task-critical regions (e.g., low-confidence or unfamiliar objects) and fewer bits to redundant background or high-confidence areas, the resulting QP map preserves machine task accuracy at substantially lower bitrates. 

\section{Methodology}
\subsection{Framework}
\begin{figure}[t]
  \centering
        \includegraphics[width=\columnwidth]{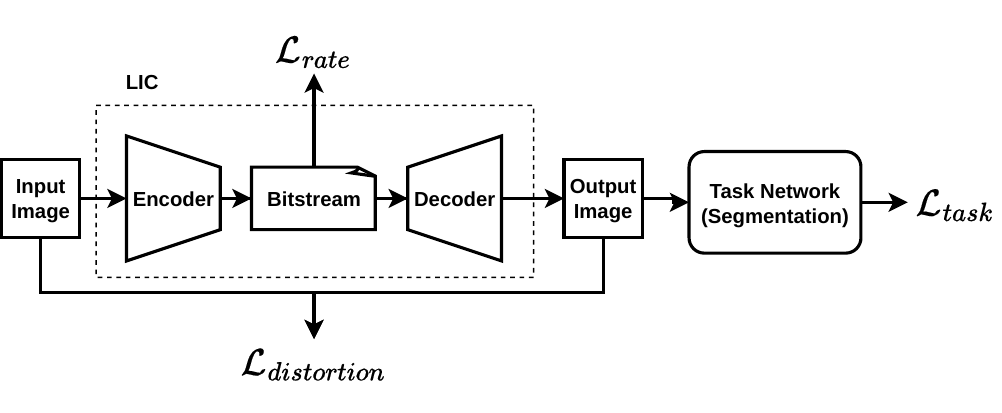}

  \caption{Joint ICM framework. The compression network is trained and the frozen segmentation network supplies the task loss.}
  \label{fig:framework}
\end{figure}

The joint compression and task model couples an LIC codec, with trainable parameters $\theta$, to a downstream task network with parameters $\phi$ (Fig.~\ref{fig:framework}). The
codec comprises a main auto-encoder---an analysis transform $g_a$ and a synthesis transform $g_s$. An input image $x$ is encoded to a quantized latent $\hat{y}$ and decoded to a reconstruction $\hat{x}$, which is passed to the task network. During training we optimize only $\theta$; the task network is frozen and provides the differentiable task loss, yet its gradients still propagate into the codec. Three losses are used: the rate $R$, defined as the negative log-likelihood of $\hat{y}$ and $\hat{z}$ under the entropy model; the distortion $D$, defined as $D=\mathrm{MSE}(x,\hat{x})$; and the task loss $T$ given as the pixel-wise cross-entropy of the segmentation prediction on $\hat{x}$.

\subsection{Baselines}
The baseline is a pure rate--distortion-based learned codec, trained for human vision, whose reconstruction is fed to the task network only at inference. We also build a naive ICM method, the Joint Optimization (JO), which is the unconstrained joint model that adds the task loss during training.
\begin{equation}
L_{\mathrm{Baseline}} = R + \lambda_1\, D, \qquad
L_{\mathrm{JO}} = R + \lambda_1 D + \lambda_2 T
\label{eq:baselines}
\end{equation}

\subsection{Quality-Constrained Formulation}
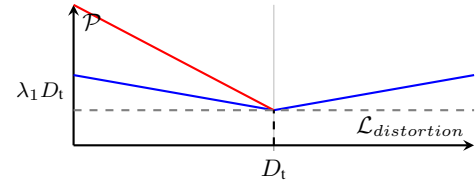
\begin{figure}[b]
\centering
\begin{tikzpicture}
\begin{axis}[
    axis lines = middle,
    xlabel = {$\mathcal{L}_{distortion}$},
    ylabel = {$\mathcal{P}$},
    xtick = {0, 5},
    xticklabels = {$0$, $D_{\text{t}}$},
    ytick = {0},
    ymin=0, ymax=10,
    xmin=0, xmax=10,
    domain=0:10,
    samples=200,
    grid=major,
    clip=false,
    thick,
    width=0.8\columnwidth,
    height=0.4\columnwidth,
]
\def\lambdone{0.5}
\def\lambdonep{1.5}
\def\Dtarget{5}
\def\constant{(\lambdone)*(\Dtarget)}
\addplot[blue, thick, domain=5:10] {\lambdone*x};
\addplot[blue, thick, domain=0:\Dtarget] {-\lambdone*(x - \Dtarget)+ \constant};
\addplot[red, thick, domain=0:\Dtarget] {-\lambdonep*(x - \Dtarget) + \constant};
\addplot[dashed, black] coordinates {(\Dtarget, 0) (\Dtarget, \constant)};
\addplot[dashed, gray] coordinates {(0, \constant) (10, \constant)};
\node[anchor=south east, font=\footnotesize] at (axis cs:0, 2.6) {$\lambda_1 D_{\text{t}}$};
\end{axis}
\end{tikzpicture}
\caption{Penalty functions w.r.t.\ distortion. \textcolor{blue}{Blue} shows absolute penalty $\lambda_1 \mathcal{L}_{distortion}$. The bilinear $\lambda_1\mathcal{P}(\mathcal{L}_{distortion},\mathcal{D}_{Target})$ in \textcolor{red}{red} applies a steeper slope after reaching the target.}
\label{fig:penalty}
\end{figure}

Rather than allowing the weighting to set the quality implicitly, we require the quality (in fact the distortion associated with the quality) to meet a user-defined target. The user specifies the target quality as a PSNR value
$\mathrm{PSNR}_{\mathrm{t}}$. This leads to the optimization in (\ref{eq:constrained}).
\begin{equation}
\min_{\theta}\; R + \lambda\, T \quad\text{s.t.}\quad  \mathrm{PSNR}(x,\hat{x}) = \mathrm{PSNR}_{\mathrm{t}} 
\label{eq:constrained}
\end{equation}

Because PSNR is a monotonic transform of MSE, the target is specified as a target PSNR and converted to a distortion target: $\mathrm{PSNR}=10\log_{10}(255^2/\mathrm{MSE})$, so $D_{\mathrm{t}}=255^2/10^{\mathrm{PSNR}_{\mathrm{t}}/10}$. Hence, the optimization problem is essentially written as
\begin{equation}
\min_{\theta}\; R + \lambda\, T \quad\text{s.t.}\quad  \mathrm{D}(x,\hat{x}) = \mathrm{D}_{\mathrm{t}} 
\label{eq:constrained_distortion}
\end{equation}
Applying the penalty method in optimization to~(\ref{eq:constrained_distortion}) converts the constrained problem into the unconstrained objective
\begin{equation}
L = R + \lambda_1\, \mathcal{P}\!\big(D(x,\hat{x}),\, D_{\mathrm{t}}\big) + \lambda_2\, T ,
\label{eq:proposed}
\end{equation}

where the penalty function $\mathcal{P}$ replaces the plain distortion term. The proposed penalties share a two-phase behavior: when the target quality has not been reached ($D\ge D_{\mathrm{t}}$) they reduce to standard rate--distortion ($\lambda_1 D$). Once the target is reached ($D<D_{\mathrm{t}}$) a penalty discourages spending further bits on visual quality.
\subsection{Penalty Functions}
The \emph{absolute} penalty is symmetric about the target: once the reconstruction exceeds the target quality it is penalized at the same rate at which sub-target distortion is rewarded,
\begin{equation}
\mathcal{P}_{\mathrm{abs}}(D,D_t) = |D - D_t| + D_t 
\label{eq:absolute}
\end{equation}

While it is effective at higher bitrates, we observed that it can underperform at low bitrates, where the penalty coefficient $\lambda_1$ is small (to meet the low bitrate) and the model still over-shoots the target. The \emph{bilinear} penalty (Fig.~\ref{fig:penalty}) remedies this by applying a steeper slope $\lambda_1'$ once the target is passed, pushing the model back toward the target instead of over-spending bits on fidelity:
\begin{equation}
\mathcal{P}_{\mathrm{bi}}(D,D_t) =
\begin{cases}
D, & D \ge D_t,\\[2pt]
-\dfrac{\lambda_1'}{\lambda_1} D + \Big(1+\dfrac{\lambda_1'}{\lambda_1}\Big) D_t, & D < D_t ,
\end{cases}
\label{eq:bilinear}
\end{equation}

While the optimization can be controlled via setting $\lambda_1'$, $\lambda_1'=\,2\lambda_1$ has proved to be a suitable value in our experiments. This tunable slope makes the bilinear penalty more flexible, and as shown in experimental results, it performs better than the absolute penalty.
\section{Results and Analysis}

\subsection{Experimental Setup}
\begin{figure}[b]
  \centering
      \includegraphics[width=0.7\linewidth]{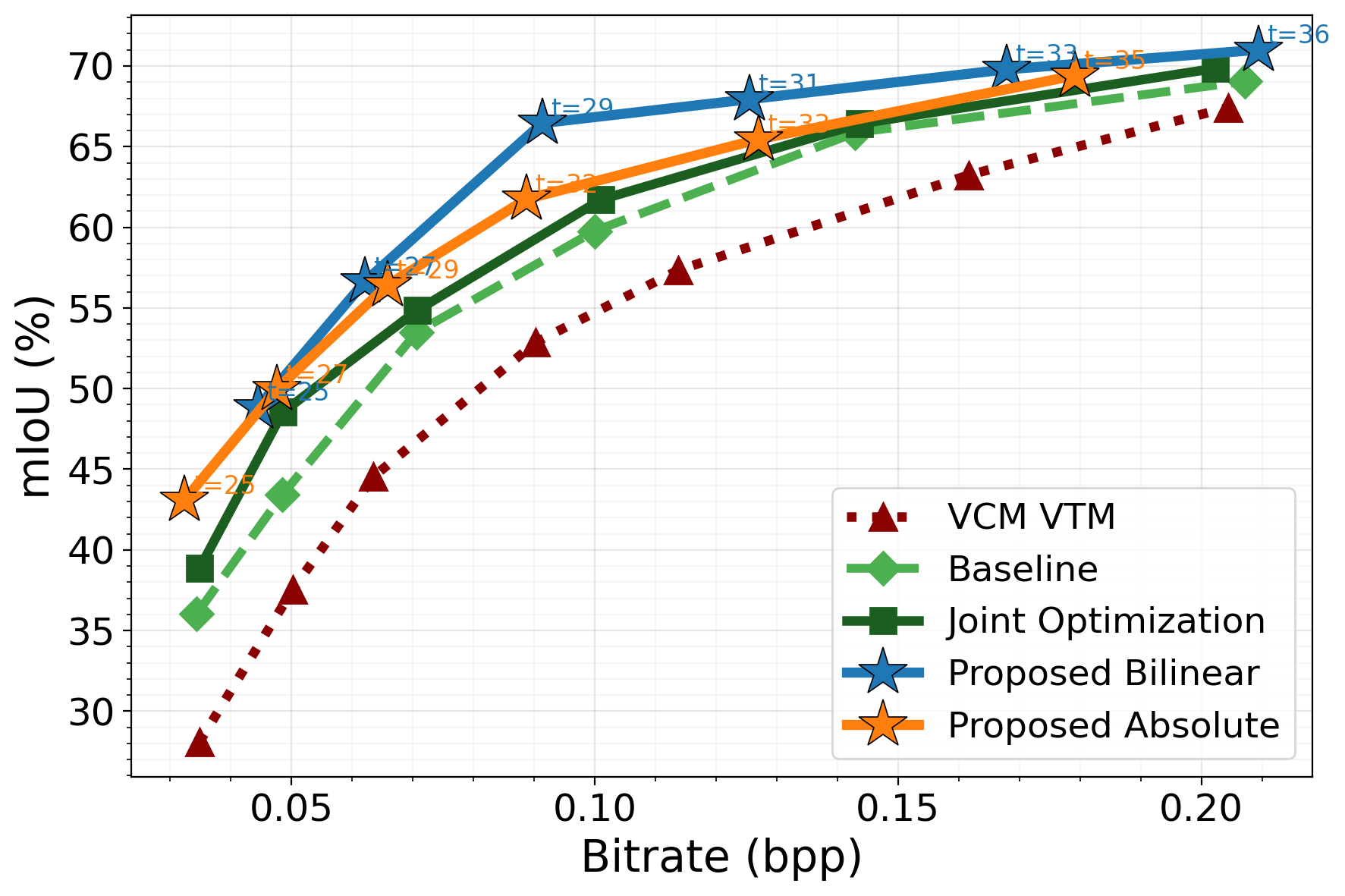}
  \caption{Rate--performance curves on Cityscapes val. The
  proposed penalties improve mIoU at equal bitrate.}
  \label{fig:curves}
\end{figure}

We evaluate on Cityscapes (2975 training and 500 validation images, 19 classes), training on $256\times256$ patches with random cropping and validating on center crops. The codec is the attention-based Cheng2020 model from CompressAI~\cite{begaintCompressAIPyTorchLibrary2020,chengLearnedImageCompression2020} across its six quality levels. The task network is Mask2Former~\cite{chengMaskedattentionMaskTransformer2022} with a Swin-T backbone~\cite{liuSwinTransformerHierarchical2021}, pretrained on Cityscapes and kept frozen. Following the CompressAI protocol we use a dual optimizer (Adam at $10^{-4}$ for the network and $10^{-3}$ for the entropy auxiliary parameters) with a plateau learning-rate schedule with a patience of $10$ epochs. We report bits per pixel (bpp), PSNR, and mIoU computed with the standard Cityscapes script \cite{cordtsCityscapesDatasetSemantic2016}, and summarize rate--task efficiency with the Bjontegaard delta rate (BD-rate) on the mIoU axis, using Baseline and Joint Optimization as the references. For Baseline, $\lambda$ $\in$ $\{$ $0.0018$, $0.0035$, $0.0067$, $0.013$, $0.025$, $0.483$ $\}$. For the JO and the proposed methods, $\lambda_1$ $\in$ $\{$$0.0018$, $0.0035$, $0.0067$, $0.013$, $0.025$, $0.483$$\}$ and $\lambda_2$ $\in$ $\{$$0.001$, $0.01$$\}$. The loss-term weights are kept constant throughout training. 

\subsection{Quantitative Results}

\begin{figure}[t]
  \centering
  \includegraphics[width=\columnwidth]{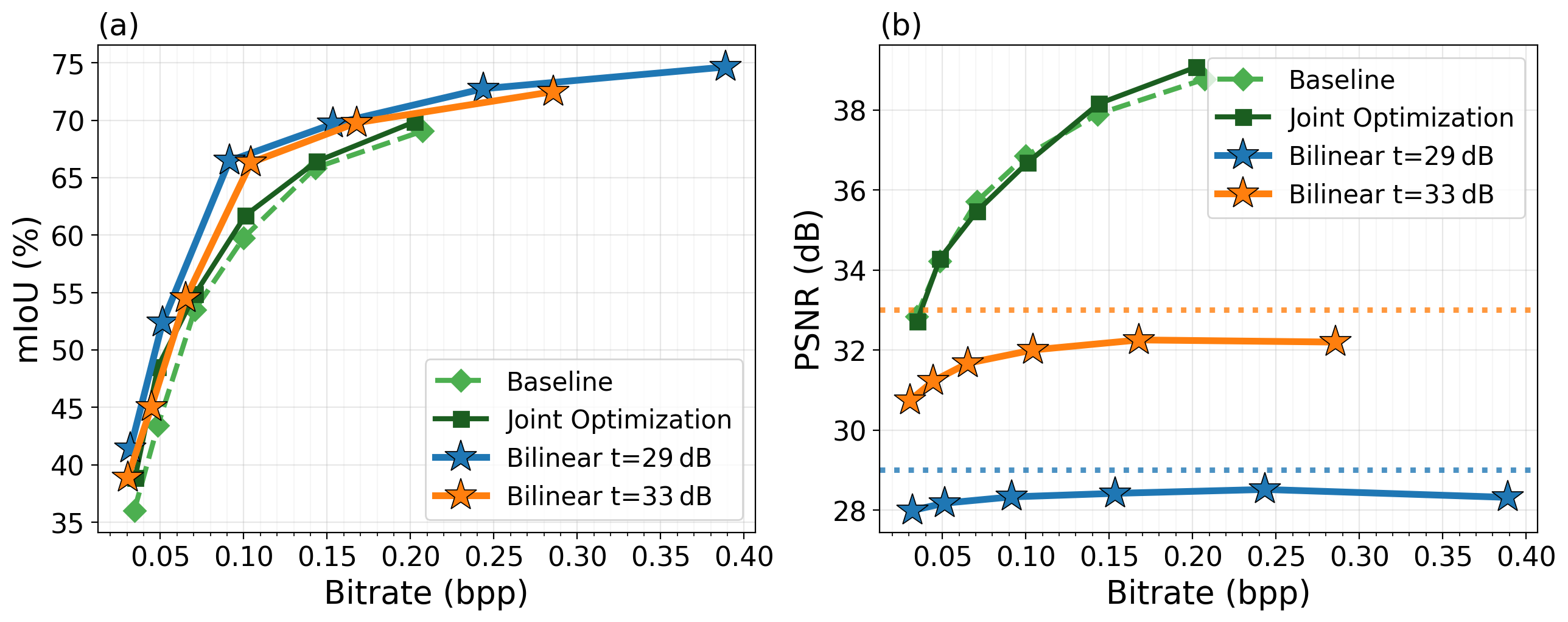}
\caption{Proposed Bilinear at PSNR targets of 29 and 33\,dB versus Baseline and Joint
         Optimization. (a)~mIoU vs bitrate: both proposed curves lie above the baselines,
         the 29\,dB target highest. (b)~PSNR vs bitrate: the constraint holds quality near
         each target (dotted), unlike the unconstrained baselines.}
  \label{fig:proposed_t29}
\end{figure}

Figure~\ref{fig:curves} reports the rate--performance behavior: both proposed penalties lie above the two baselines at equal bitrate, with the largest margin in the low-to-mid bitrate range, and the bilinear penalty performs best throughout. Both penalties also outperform the standard codec VTM, used as the anchor in MPEG VCM \cite{gaoRecentStandardDevelopment2021}. Since the useful target quality grows with the bitrate, we further adopt a mixed-target schedule that sets a lower target in the low-bitrate region than in the high-bitrate region, as shown in Figure~\ref{fig:curves}. Table~\ref{tab:bdrate} summarizes the BD-rate based on mIoU. At equal task performance, the bilinear penalty achieves a BD-rate of $-22.82\%$ over joint optimization and $-29.81\%$ over the Baseline. We also analyze each target individually, fixing the target quality across all bitrates. Table~\ref{tab:bdrate_per_psnr} reports the BD-rate by target PSNR level and confirms that a bitrate saving can be achieved for a wide range of target qualities. Figure~\ref{fig:proposed_t29} presents two examples of fixed quality: both mIoU curves show an advantage over the baselines, and the constraint is reflected in the PSNR curves being held near the target quality.

\begin{table}[t]
  \centering
  \caption{BD-Rate (\%) and BD-mIoU relative to reference curves. Negative BD-Rate or positive BD-mIoU point to improvement.}
  \label{tab:bdrate}
  \begin{tabular}{llrr}
    \toprule
    Proposed & Reference & BD-Rate (\%) & BD-mIoU(\%) \\
    \midrule
    \multirow{3}{*}{Bilinear} & Baseline            & -29.81 & 5.60 \\
                              & Joint Optimization  & -22.82 & 3.86 \\
                              & VCM VTM             & -45.78 & 10.75 \\
    \midrule
    \multirow{3}{*}{Absolute} & Baseline            & -20.69 & 4.29 \\
                              & Joint Optimization  & -11.32 & 2.13 \\
                              & VCM VTM             & -38.44 & 10.29 \\
    \bottomrule
  \end{tabular}
\end{table}

\begin{table}[t]
  \renewcommand{\arraystretch}{1}
  \caption{BD-Rate (\%) per target-PSNR level on Cityscapes val (quality axis = mIoU).
           Each proposed curve uses the best $\lambda_2$ per quality point.
           BL = Baseline, JO = Joint Optimization.}
  \label{tab:bdrate_per_psnr}
  \centering
  \setlength{\tabcolsep}{3.5pt}
  \begin{tabular}{c rr rr}
    \toprule
    & \multicolumn{2}{c}{Proposed Bilinear} & \multicolumn{2}{c}{Proposed Absolute} \\
    \cmidrule(lr){2-3}\cmidrule(lr){4-5}
    Target (dB) & vs BL & vs JO & vs BL & vs JO \\
    \midrule
    25 & $-14.6$ & $-5.9$  & $-20.8$ & $-11.0$ \\
    26 & $-15.3$ & $-5.6$  & $-19.2$ & $-8.9$  \\
    27 & $-21.1$ & $-11.6$ & $-20.5$ & $-10.3$ \\
    28 & $-21.1$ & $-11.4$ & $-23.5$ & $-14.2$ \\
    29 & $-30.8$ & $-22.5$ & $-22.4$ & $-12.7$ \\
    30 & $-28.3$ & $-19.8$ & $-18.7$ & $-9.4$  \\
    31 & $-24.5$ & $-15.3$ & $-10.5$ & $-0.4$  \\
    32 & $-17.6$ & $-7.5$  & $-18.9$ & $-9.4$  \\
    33 & $-20.0$ & $-10.1$ & $-8.7$  & $+1.8$  \\
    34 & $-14.2$ & $-3.9$  & $-8.2$  & $+2.9$  \\
    35 & $-12.8$ & $-3.0$  & $-7.5$  & $+3.9$  \\
    36 & $-15.9$ & $-5.7$  & $-6.9$  & $+4.4$  \\
    \bottomrule
  \end{tabular}
\end{table}

\subsection{Analysis}
\begin{figure*}[t]
  \centering
  \subfloat[Original]{%
    \begin{tikzpicture}[spy using outlines={rectangle, red, very thick,
        magnification=4, size=1.2cm, connect spies}]
      \node[inner sep=0] (img) {\includegraphics[width=0.24\textwidth]{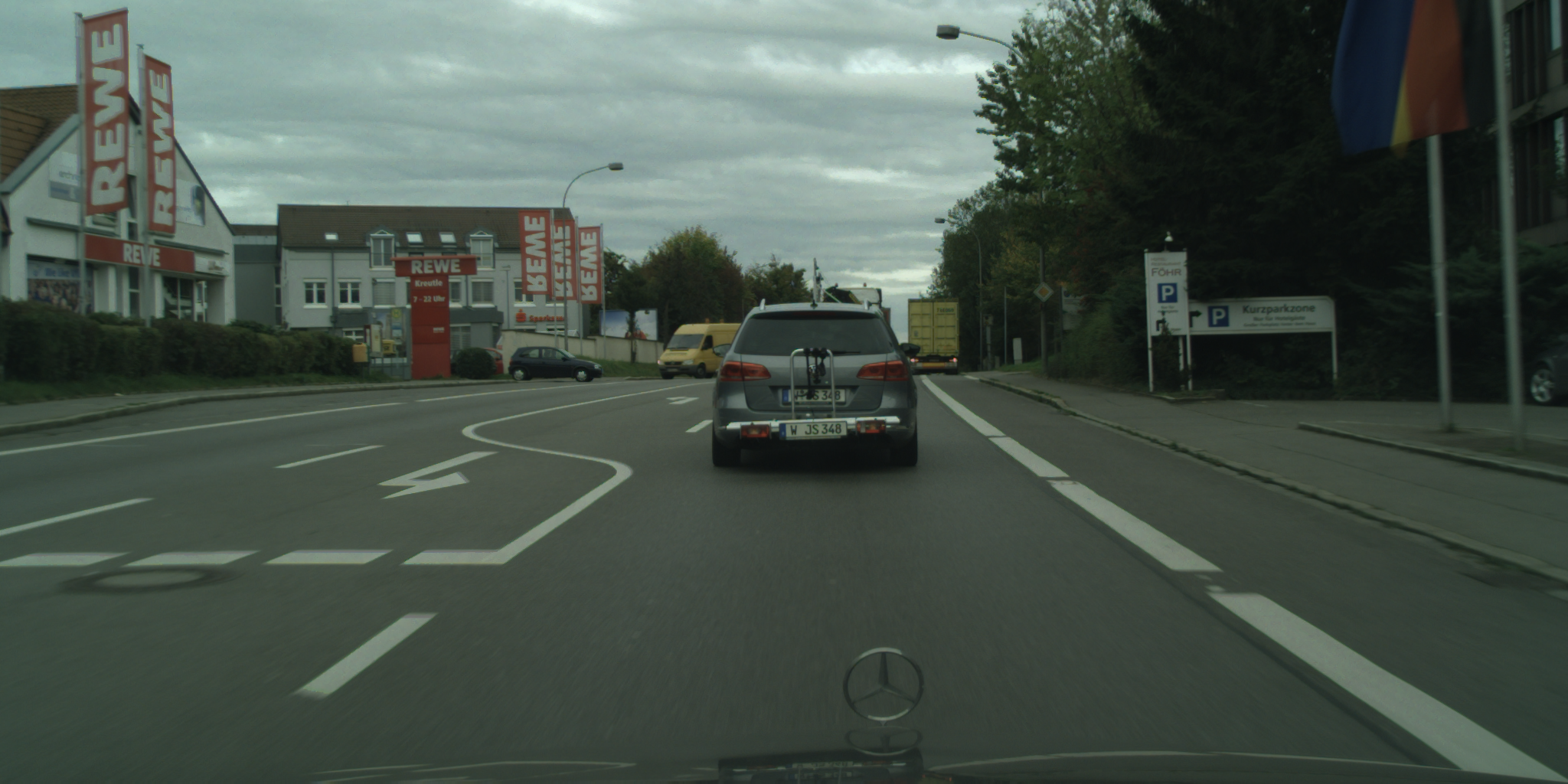}};
      \spy on (0,0) in node [anchor=south east] at (img.south east);
    \end{tikzpicture}}\hfill
  \subfloat[Baseline bpp=0.0301 \\
  PSNR=36.49 dB, mIoU=33.2\%]{%
    \begin{tikzpicture}[spy using outlines={rectangle, red, very thick,
        magnification=4, size=1.2cm, connect spies}]
      \node[inner sep=0] (img) {\includegraphics[width=0.24\textwidth]{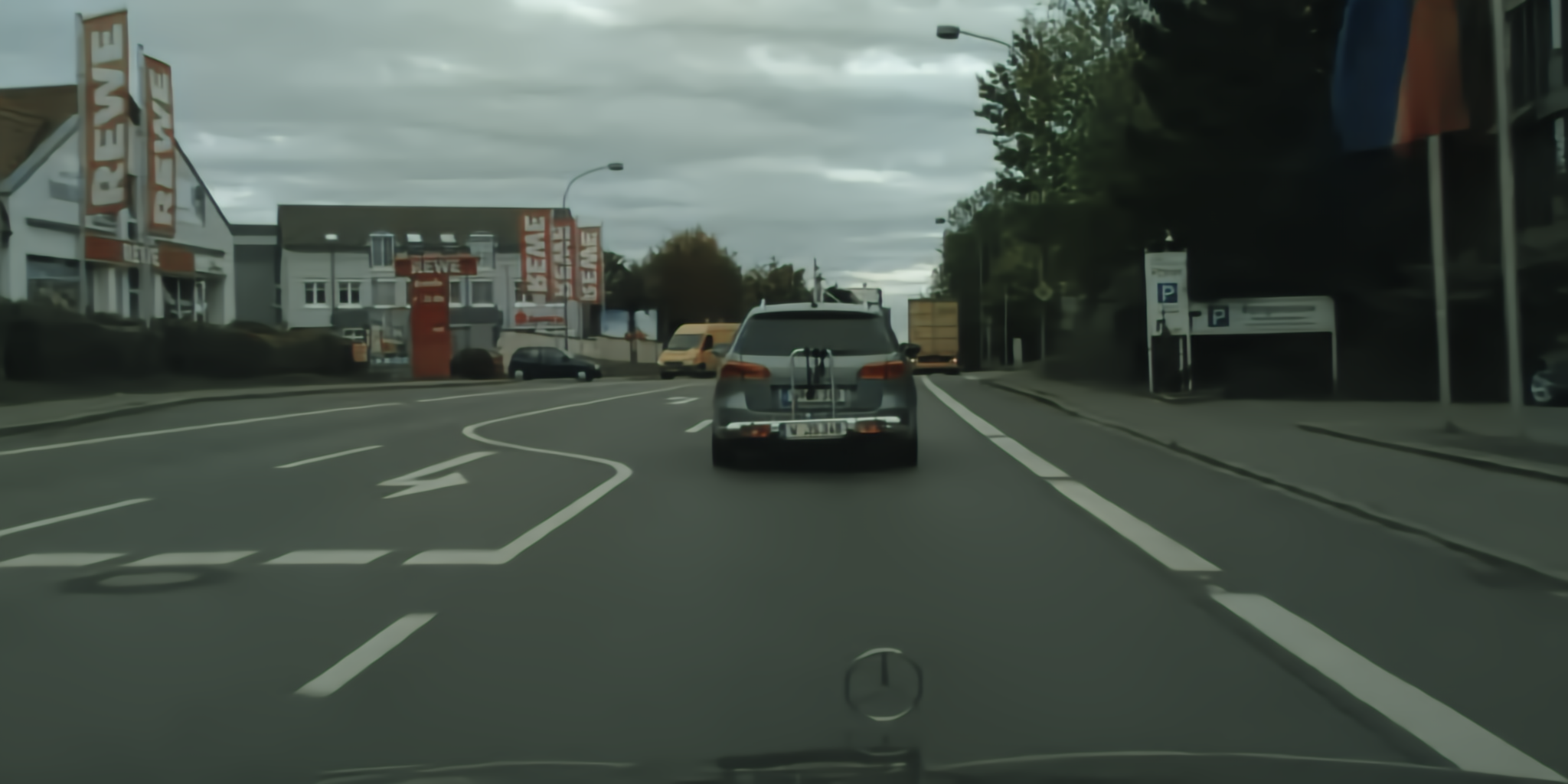}};
      \spy on (0,0) in node [anchor=south east] at (img.south east);
    \end{tikzpicture}}\hfill
  \subfloat[JO bpp=0.0302  \\
  PSNR=36.77 dB  mIoU=38.1\%]{%
    \begin{tikzpicture}[spy using outlines={rectangle, red, very thick,
        magnification=4, size=1.2cm, connect spies}]
      \node[inner sep=0] (img) {\includegraphics[width=0.24\textwidth]{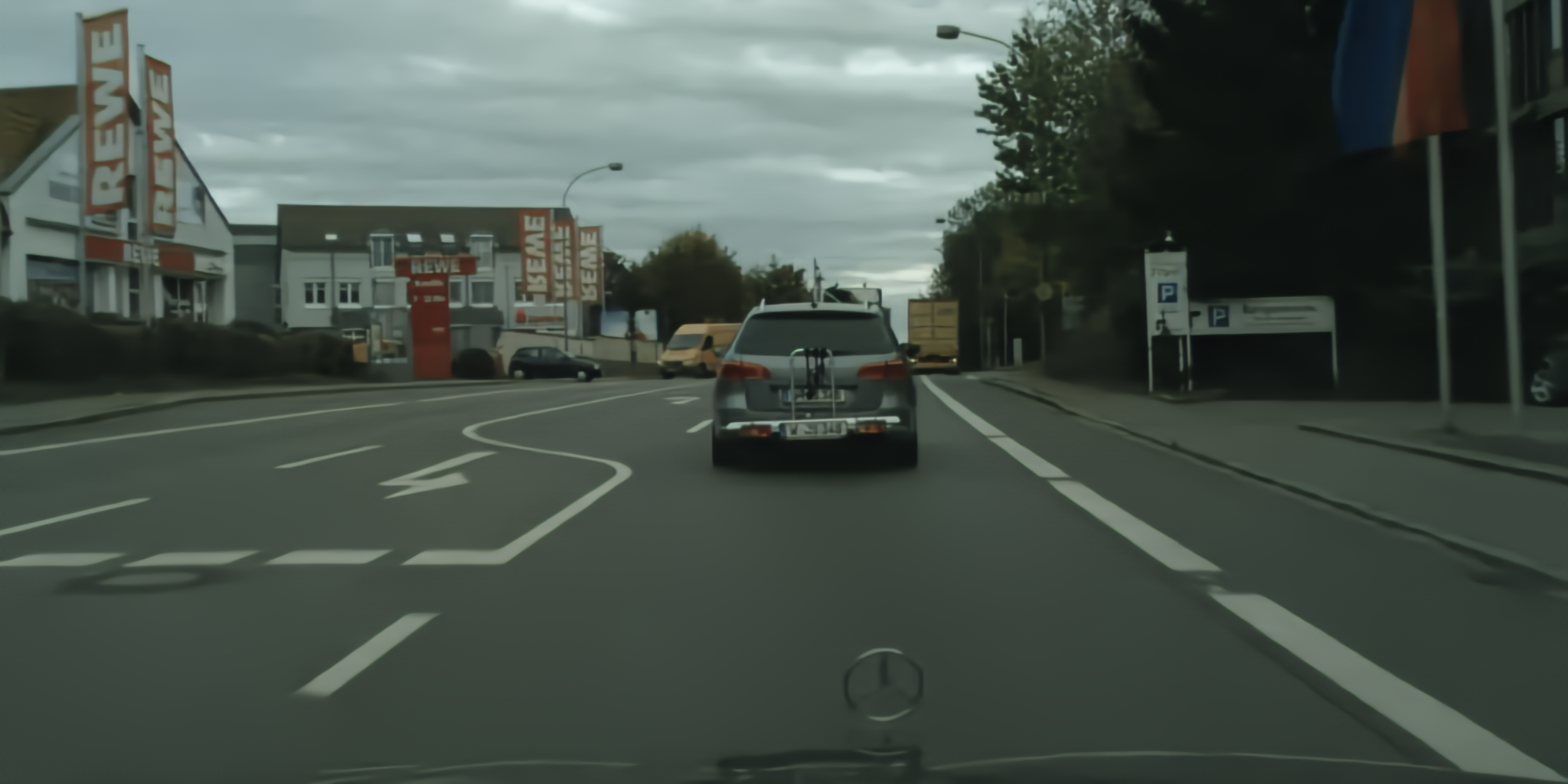}};
      \spy on (0,0) in node [anchor=south east] at (img.south east);
    \end{tikzpicture}}\hfill
  \subfloat[Proposed bpp=0.0300  \\
  PSNR=31.06 dB  mIoU=42.4\%]{%
    \begin{tikzpicture}[spy using outlines={rectangle, red, very thick,
        magnification=4, size=1.2cm, connect spies}]
      \node[inner sep=0] (img) {\includegraphics[width=0.24\textwidth]{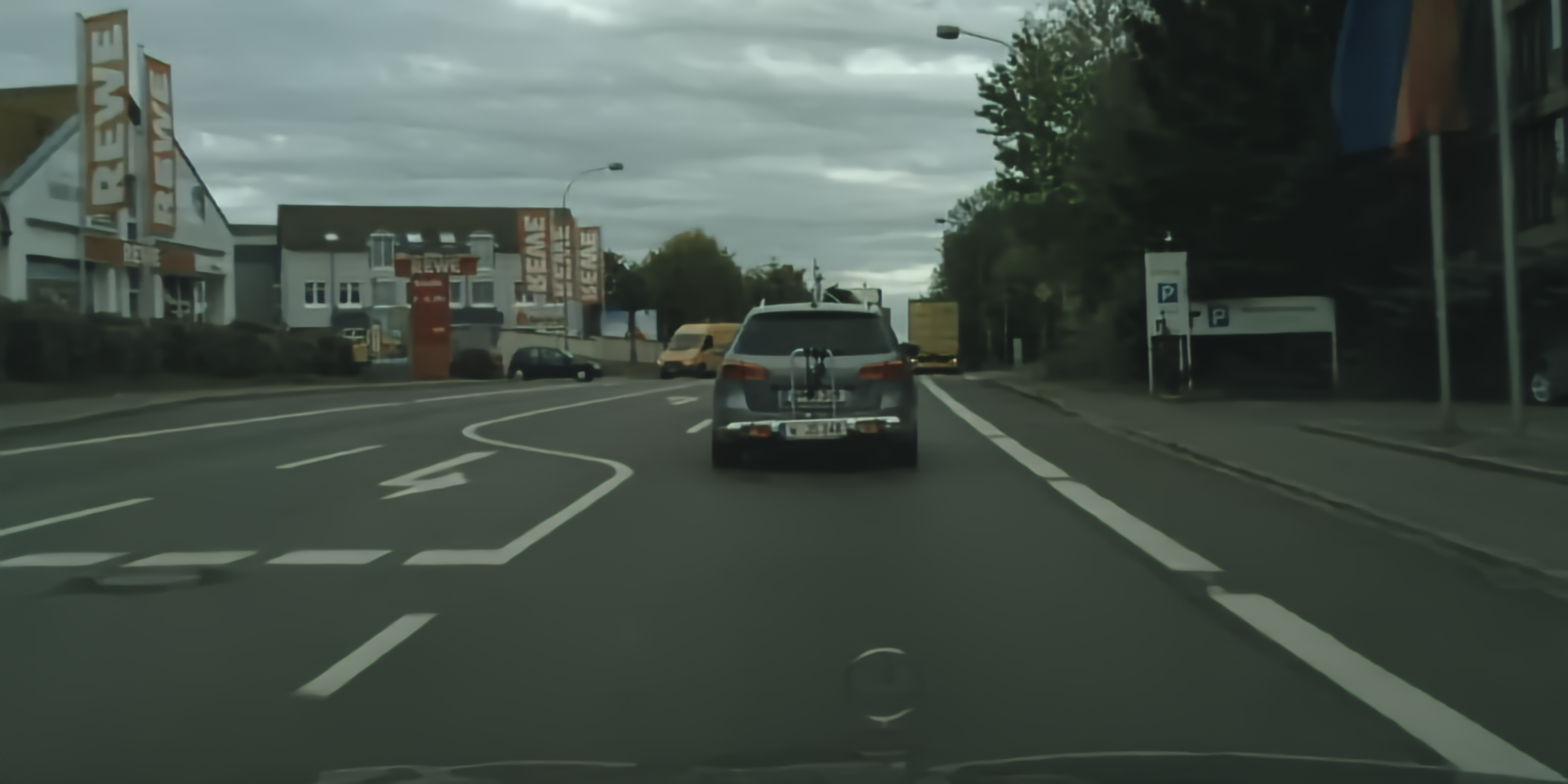}};
      \spy on (0,0) in node [anchor=south east] at (img.south east);
    \end{tikzpicture}}\\[-8pt]
  \subfloat[Bits: Proposed $-$ Baseline]{\includegraphics[width=0.24\linewidth]{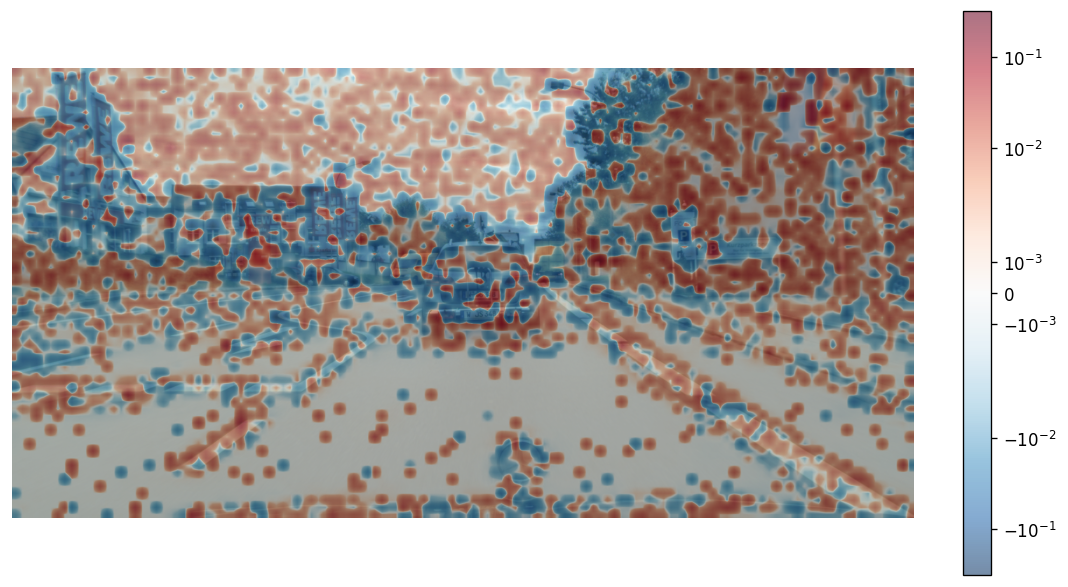}}\hspace{6pt}
  \subfloat[Bits: Proposed $-$ JO]{\includegraphics[width=0.24\linewidth]{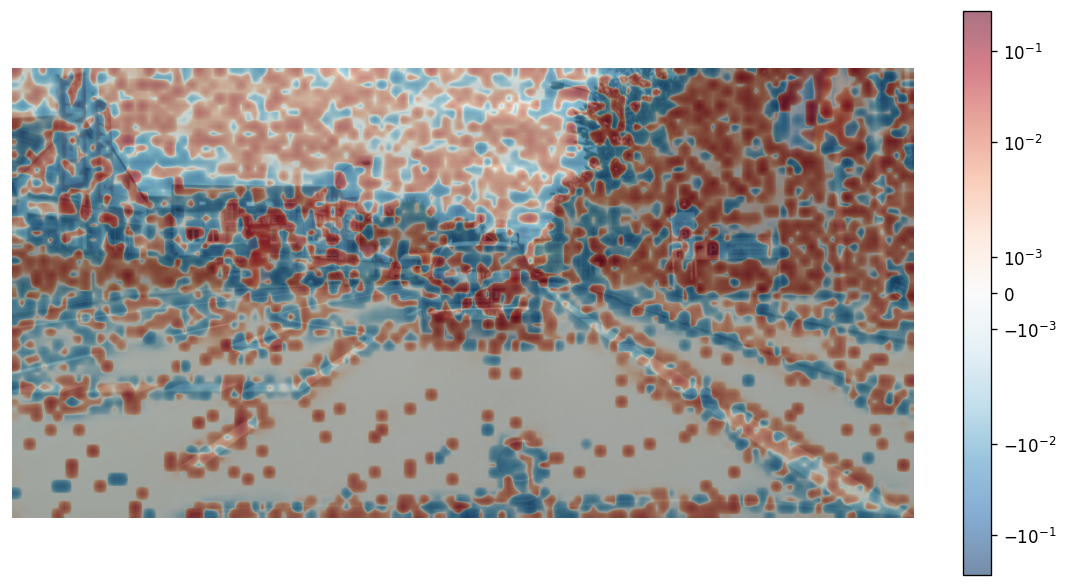}}\hspace{6pt}
  \subfloat[Segmentation Win/Lose]{\includegraphics[width=0.24\linewidth]{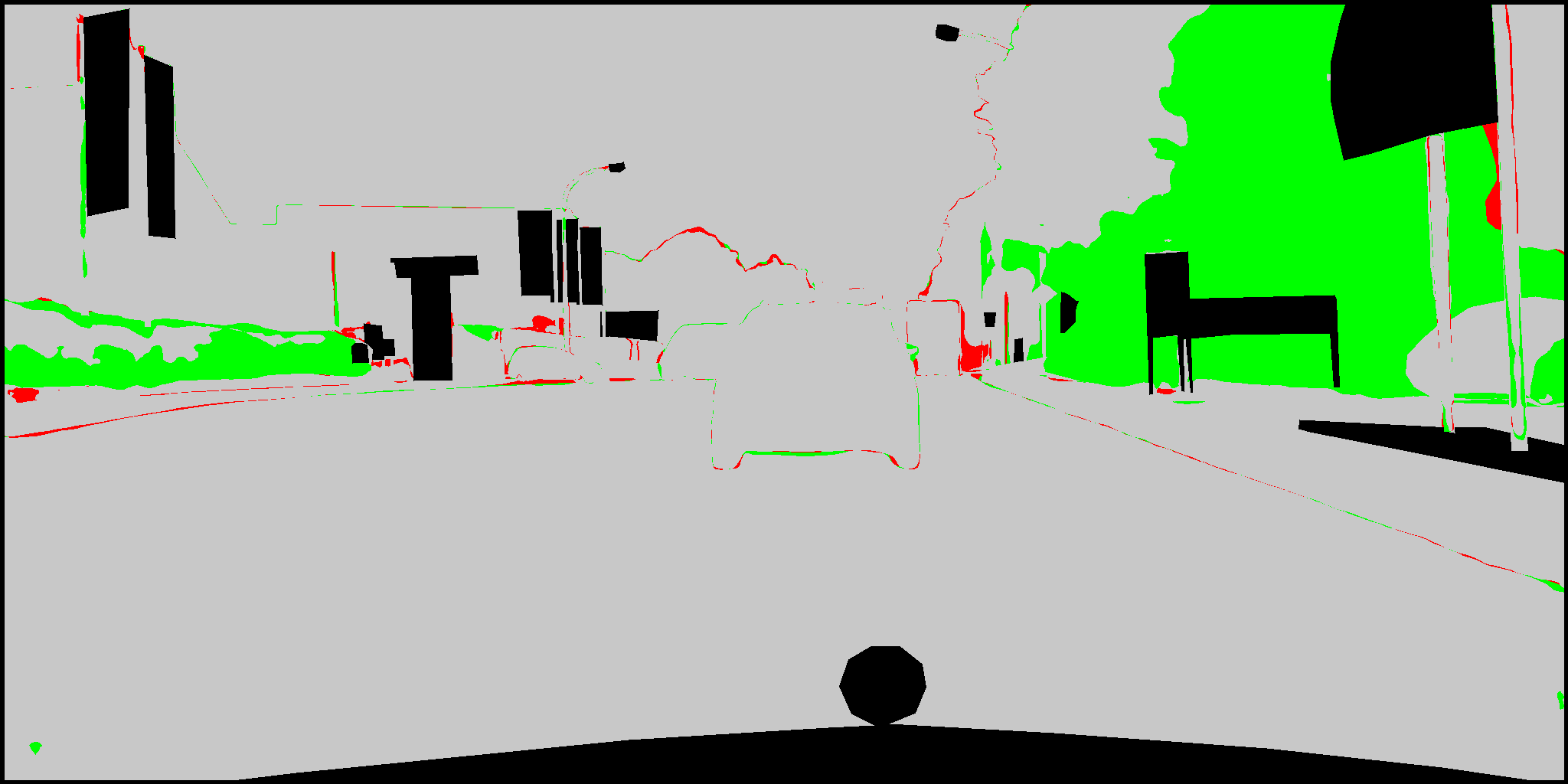}}\\[-8pt]
  \subfloat[Latent: Baseline]{\includegraphics[width=0.24\linewidth]{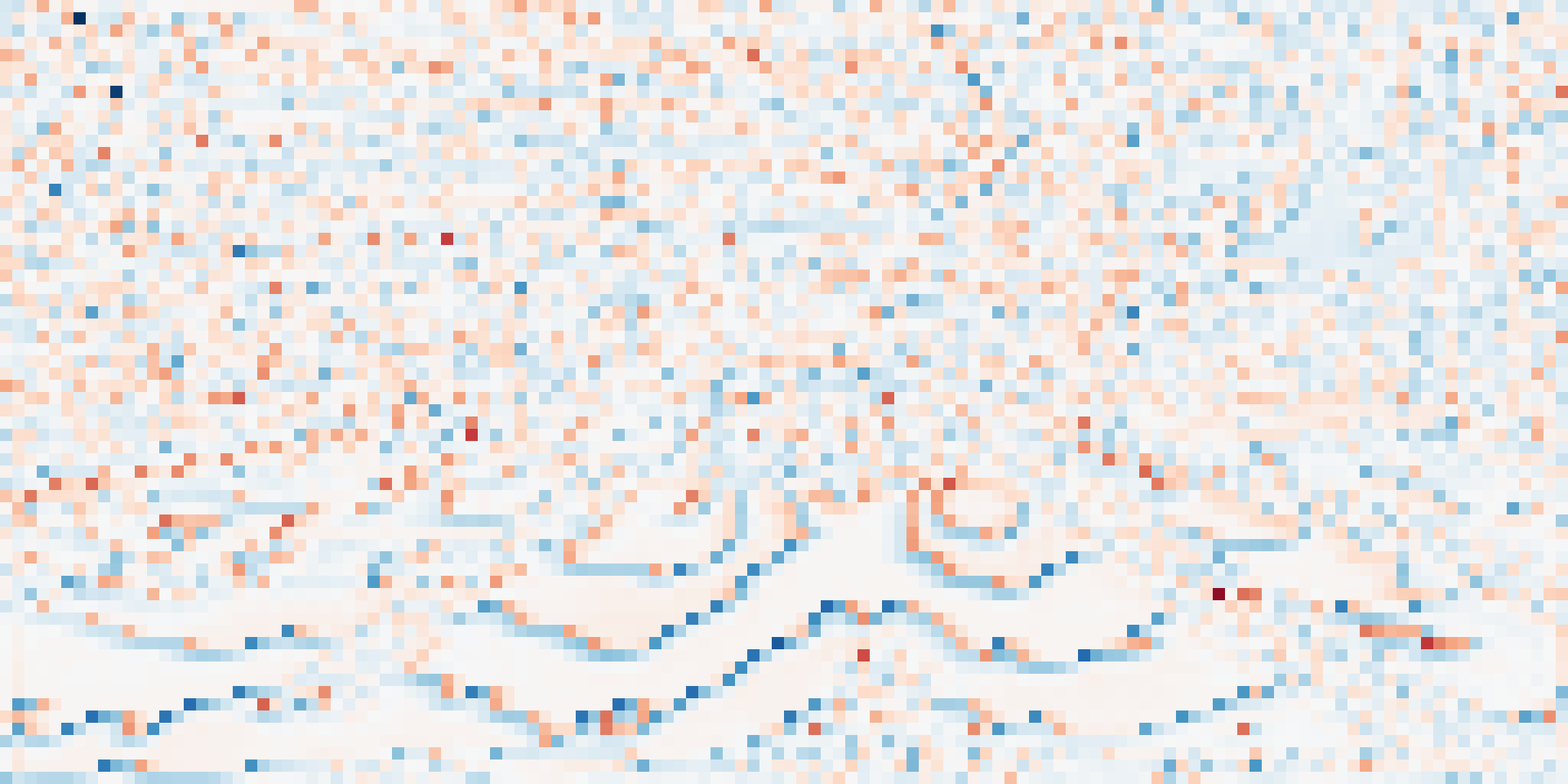}}\hspace{6pt}
  \subfloat[Latent: JO]{\includegraphics[width=0.24\linewidth]{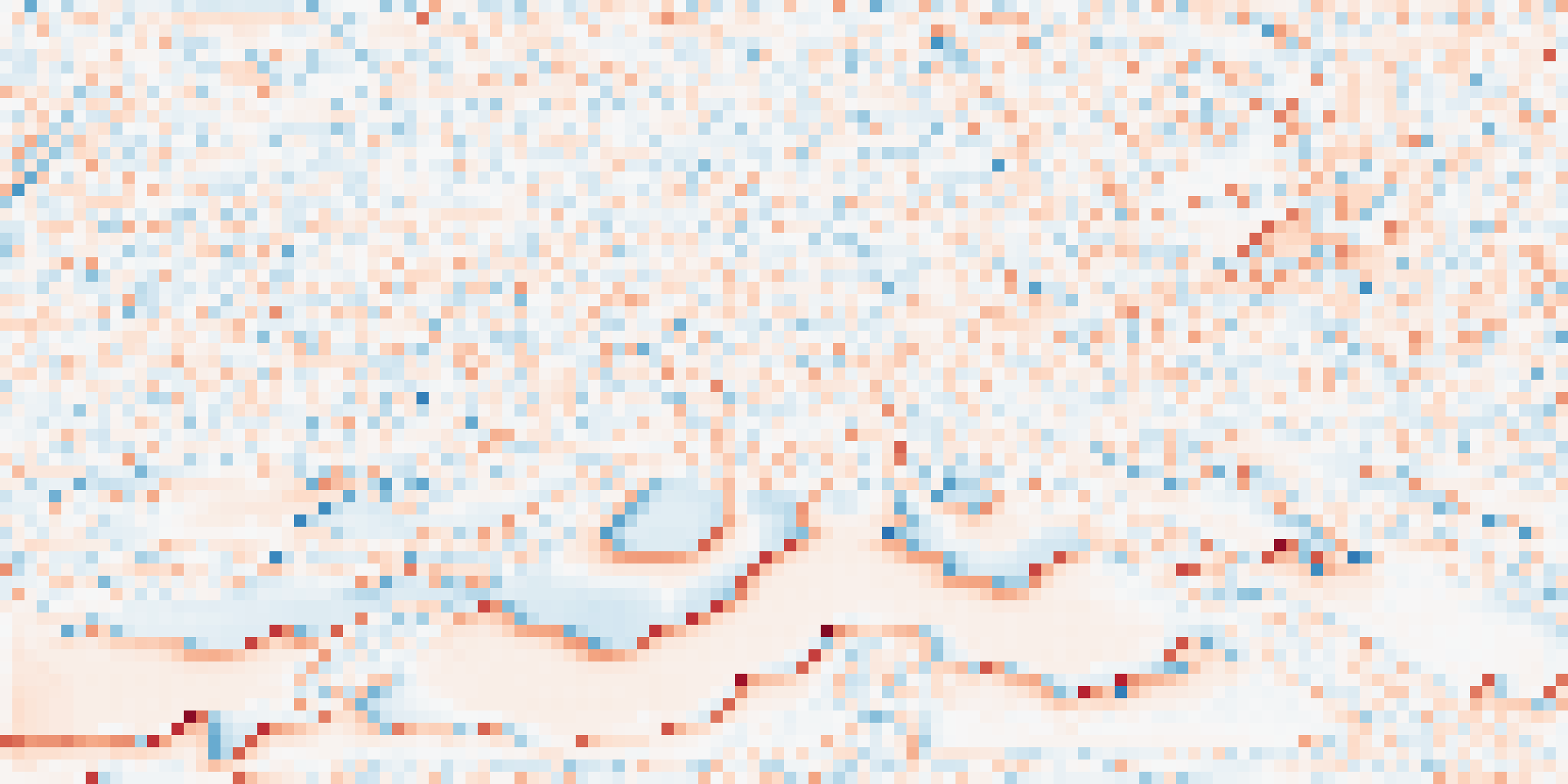}}\hspace{6pt}
  \subfloat[Latent: Proposed]{\includegraphics[width=0.24\linewidth]{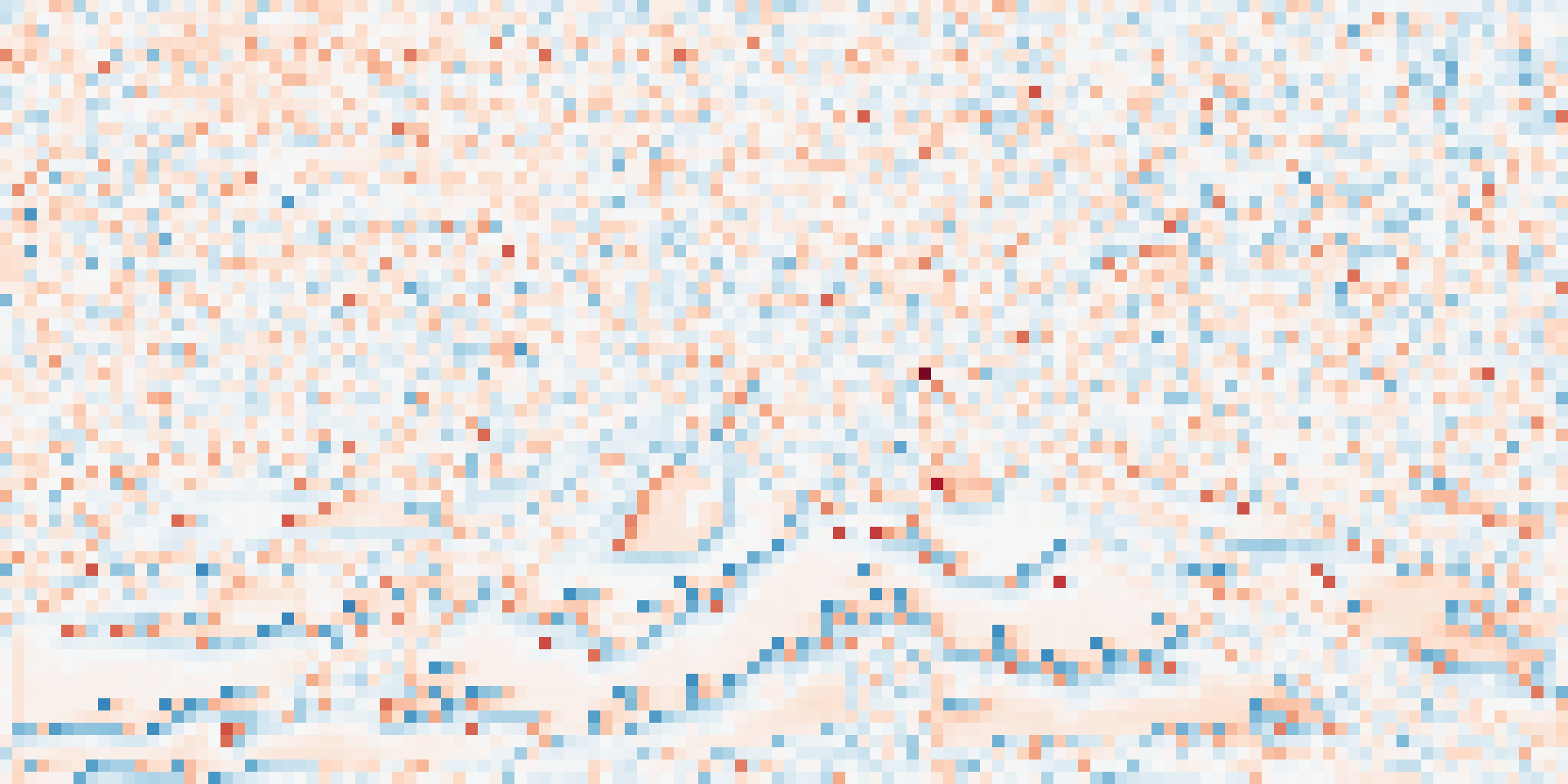}}
  \caption{Analysis of lindau\_000003\_000019 at equal bitrate ($\approx$0.030~bpp).
  \textbf{Top:} reconstructions; the proposed method trades $\sim$5~dB of PSNR for
  $+$9.2 points of mIoU. \textbf{Middle:} per-pixel bit difference vs.\ Baseline and
  vs.\ JO (red~=~more bits, blue~=~fewer), and proposed-vs-JO segmentation
  (red/green~=~only JO/only proposed correct, gray~=~tie).
  \textbf{Bottom:} normalized latent channel with highest bitrate.}
  \label{fig:qual}
\end{figure*}
To understand \emph{why} the constraint helps, we inspect and compare reconstructed images of Baseline, unconstrained Joint Optimization (JO), and the Proposed bilinear penalty, at a similar bitrate. We present one representative low-bitrate scene (lindau\_000003\_000019, $\approx$0.030~bpp) in this section. Analyzing low-bitrate examples helps to understand how the proposed method works under a tight resource constraint.

\textbf{Reconstruction quality.} As observed in the top row of Fig.~\ref{fig:qual}, all methods achieve a comparable visual quality at nearly identical bitrates ($\approx$0.030~bpp). However, the proposed method deliberately loses 5~dB of PSNR (31.1 vs.\ 36.5~dB), which buys a 9.2\% improvement in segmentation performance (42.4\% vs. 33.2\%). Perceptually, the reconstruction remains acceptable: the proposed method spreads its extra error more uniformly over smooth, texture-poor areas (see the zoom insets), so the degradation lands where it matters least for segmentation while object boundaries stay sharp.

\textbf{Spatial bit allocation.} Summing the latent bit cost over channels and mapping it back to image space reveals which regions gain or lose bits under the quality constraint. The middle row of Fig.~\ref{fig:qual} shows the per-pixel bit difference of the Proposed codec relative to the Baseline and to JO: red marks added bits, blue marks savings. Bits move toward task-salient structure---vehicles, poles, and object boundaries---and away from large uniform regions such as road and sky. The segmentation error maps in the same row confirm the payoff: relative to JO, the Proposed codec recovers foreground regions that JO mislabels, consistent with its higher mIoU at equal rate.

\textbf{Latent representation.}
The normalized latent, obtained by standardizing the latent with the mean and scale predicted by the entropy model, represents the information transmitted through the bottleneck of the framework in Fig.~\ref{fig:framework}. If the entropy model predicted the latent perfectly, the normalized latent would show no spatial structure; any remaining structure reflects content the prior failed to anticipate, which costs additional bits. To identify the channels that carry most of the information, we compute the bitrate of each channel and visualize the normalized latent of the channels with the highest bitrate. As shown in Fig.~\ref{fig:qual}, in the low-bitrate regime the normalized latent of the proposed codec exhibits less spatial structure overall than those of the baselines, indicating a better-calibrated entropy model, while its bit budget is concentrated in a few high-bitrate channels whose structure follows task-relevant content such as object boundaries (bottom row of Fig.~\ref{fig:qual}). In contrast, the Baseline and JO codecs spend excessive bits on regions that should be easily predictable, such as road surfaces.
\section{Conclusion}
We presented a quality-constrained formulation for jointly training an image codec with a semantic segmentation task, which replaces the unconstrained distortion term with a penalty that enforces a target PSNR, letting the codec redirect the bits freed from perceptual fidelity toward the computer vision task. On Cityscapes the approach yields consistent BD-rate savings on the mIoU axis over both a pure rate--distortion baseline and an unconstrained joint optimization scheme, while holding the reconstruction close to the requested quality. Our analyses show that the codec shifts bits toward task-relevant regions. Two findings stand out: targeting a specific distortion level in coding for machines is feasible, and a moderate penalty enforcement improves task accuracy. Future work may dynamically adapt the target quality spatially, using JND as the enforcement threshold so that bits are allocated according to local regional sensitivity.
\clearpage
\bibliographystyle{IEEEbib}
\bibliography{references}

\clearpage

\end{document}